SIM-Net: A Multimodal Fusion Network Using Inferred 3D Object Shape Point Clouds from RGB Images for 2D Classification

*Youcef Sklab*[1*], *Hanane Ariouat*[1], *Eric Chenin*[1], *Edi Prifti*[1,2], Jean-*Daniel Zucker*[1,2]

[1]IRD, Sorbonne Université, UMMISCO, Bondy, F-93143, France
[2]INSERM, Nutrition et Obesities; systemic approaches, NutriOmique, AP-HP, Hôpital Pitié-Salpêtrière, Paris, France
*Corresponding author, youcef.sklab@ird.fr

**abstract**
**We introduce the Shape-Image Multimodal Network (SIM-Net), a novel 2D image classification architecture that integrates 3D point cloud representations inferred directly from RGB images. Our key contribution lies in a pixel-to-point transformation that converts 2D object masks into 3D point clouds, enabling the fusion of texture-based and geometric features for enhanced classification performance. SIM-Net is particularly well-suited for the classification of digitized herbarium specimens—a task made challenging by heterogeneous backgrounds, non-plant elements, and occlusions that compromise conventional image-based models. To address these issues, SIM-Net employs a segmentation-based preprocessing step to extract object masks prior to 3D point cloud generation. The architecture comprises a CNN encoder for 2D image features and a PointNet-based encoder for geometric features, which are fused into a unified latent space. Experimental evaluations on herbarium datasets demonstrate that SIM-Net consistently outperforms ResNet101, achieving gains of up to 9.9% in accuracy and 12.3% in F-score. It also surpasses several transformer-based state-of-the-art architectures, highlighting the benefits of incorporating 3D structural reasoning into 2D image classification tasks.**

1. **Introduction**

Over the past two decades, deep learning has made profound advancements, leading to the development of numerous innovative architectures that have significantly advanced the field of artificial intelligence [1,2,3,4,5,6,7,8,9,10]. In particular, the field of computer vision has seen the emergence of many high-performing models, especially for image classification [1,2,3,4,5,10]. However, traditional classification methods often struggle with complex backgrounds, occlusions, and noisy environments–challenges that are especially pronounced in biological datasets such as digitized herbarium specimens[1], which have become vital resources for botany and biodiversity research [11,12,13,14,15,16,17]. Models such as ResNet [5] and Vision Transformers [10], exhibit varying degrees of robustness to noise and differences in foreground and background sensitivity [18,19].

Analyzing digitized herbarium specimens remains challenging due to varying textures and presence of numerous non-plant elements, such us scale bars, labels, and extraneous artifacts (*cf.* Figure 1). These elements, often scattered randomly across herbarium sheets, can obscure crucial information within the specimens.

Further complicating the task, the quality of the paper used for herbarium sheets varies significantly, with some fixed-on newspapers and others on paper that has degraded and darkened over time, often to a hue like that of the plants themselves. This degradation makes it increasingly difficult for image classification models to distinguish the specimen from the background. Noise, in this context, refers to any irrelevant or extraneous information that obscures or distorts the herbarium specimen. This, in turn, complicates the task of feature classification.

---

[1] Natural history collections can preserve several hundred years of information on the evolution of biodiversity and the environment. Interest in these collections is on the rise, as they offer valuable insights for understanding and alleviating severe threats to biodiversity.

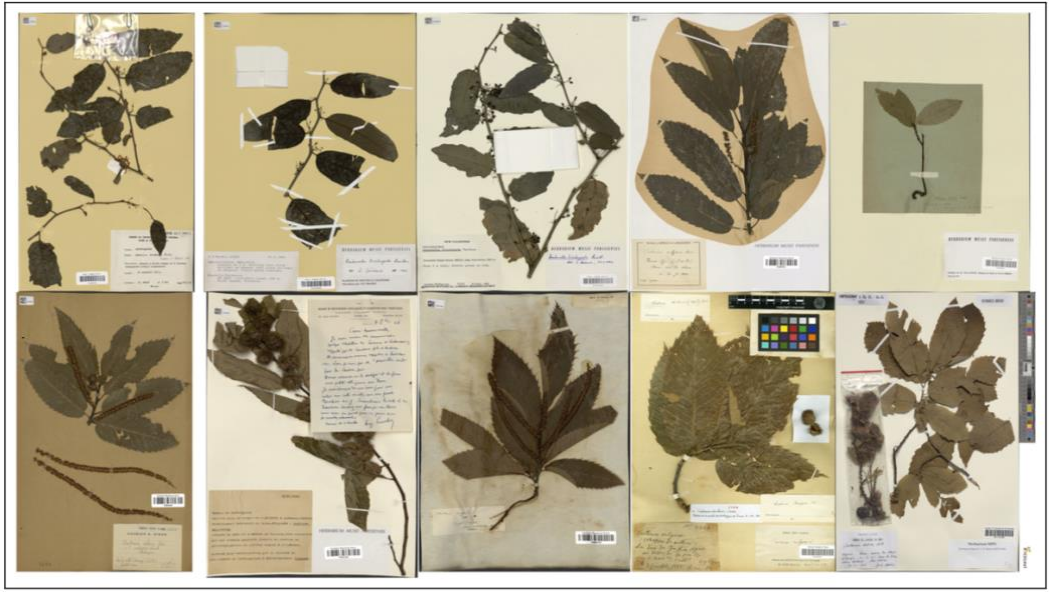

*Figure 1: Examples illustrating the diversity in paper color, quality and the non-plant elements in herbarium sheets (non-exhaustive).*

In a previous study [16], we explored herbarium image classification for detecting the presence of thorns on plants using ResNet on two datasets: one with raw images and another with segmented images where all background elements were removed, leaving only the plant. We found that while ResNet achieved better training and validation performance on the raw image dataset (95% accuracy versus 94% on the segmented dataset), the model trained on segmented images demonstrated superior generalization. When tested on a set of 60 new images (30 with and 30 without thorns), the accuracy of the ResNet model trained on raw images dropped to 28.8%. In contrast, the model trained on segmented images showed a smaller decline, achieving 58.8% accuracy. These results suggest that removing background noise enabled the model to focus more effectively on relevant plant features, thereby improving its ability to generalize to unseen data. Figure 2 illustrates the decisive regions identified by ResNet in unsegmented and segmented images. The top example shows an incorrect prediction where the model focused on the background instead of the plant. The bottom example demonstrates a correct prediction, with the model properly focusing on the plant itself. This contrast highlights the value of background removal in enhancing a model's ability to identify key features, emphasizing the importance of targeted focus in deep-learning-based plant classification. Our study highlights the critical role of dataset preprocessing in developing models with improved generalization for botanical image classification.

Although dataset preprocessing is crucial for improving classification, traditional 2D image classification models face a more fundamental limitation: they rely solely on color, texture, and spatial patterns. This reliance makes them highly sensitive to background noise and occlusions, reducing their ability to generalize in complex settings. A promising solution lies in multimodal learning, which integrates complementary data representations to enhance classification. Multimodal architectures have transformed many AI fields, particularly in 3D object detection and vision-language models [20], as demonstrated by pioneering models like CLIP [9]. In image classification, combining 2D texture information with 3D shape representations can yield richer features and greater robustness to background variations. Fusing 2D and 3D information to enhance accuracy has thus become an increasingly active area of research [21].

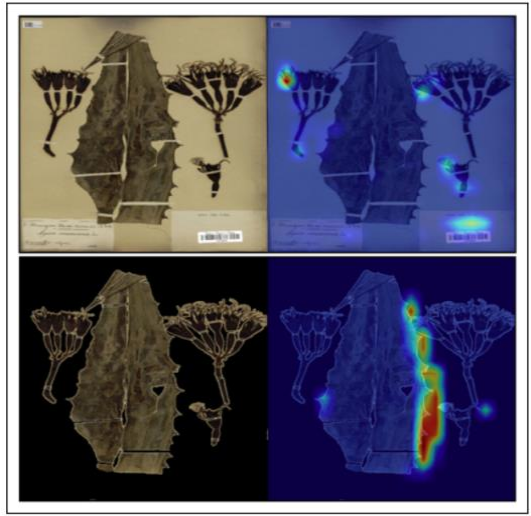

*Figure 2: The decisive parts in prediction on unsegmented and segmented images.*

This paper introduces SIM-Net —Shape-Image Multimodal Network— (Figure 3), a novel approach for 2D image classification that incorporates 3D object shape information extracted from 2D images. Our approach employs a multimodal fusion strategy, combining standard 2D images with a structured 3D representation of object shapes in the form of point clouds. These point clouds preserve essential properties such as an unordered, irregular format and permutation invariance. Unlike traditional classification models that rely solely on 2D textures and spatial arrangements within images, SIM-Net integrates geometric and structural information by transforming segmented 2D object masks into 3D point clouds. We begin by generating masks of the targeted objects–in this case, plant specimens–using segmentation techniques, and subsequently transform these masks into 3D object shape point clouds. Specifically, a point cloud is derived from an image through a pixel-to-point transformation process: each pixel, initially defined by its position coordinates (x, y) and color values (r, g, b) in the RGB color space, is converted into a 3D point. This transformation maps the pixel's spatial and color information into three coordinates (x,y,z) in the point cloud, effectively translating 2D image data into a 3D representation. The resulting pairs of 2D images and corresponding 3D point clouds are processed through a multimodal fusion architecture (SIMI-Net) composed of two networks: a CNN network (ResNet [5]) and a PointNet network [7]. The 3D shape representations provide geometric and spatial information not readily discernible in 2D images, thereby enriching the feature space available to the learning model. The 3D point cloud representation enables the network to learn plant morphology independently of background noise, lighting variations, and paper degradation—key factors that hinder the performance of purely 2D-based classifiers.

ResNet [5], with its deep convolutional layers, excels in extracting hierarchical and detailed features from 2D images, such as textures, colors, and edges. However, its capabilities are limited to the information present in two dimensions. PointNet [7] processes 3D point clouds, capturing the spatial geometry of objects. By understanding the arrangement and relationship of points in the three-dimensional space, it can glean insights about the shape and structure of objects, aspects that 2D images alone cannot provide. When these two networks are fused in a merging architecture, they complement each other, creating a more robust and comprehensive feature set for classification.

Unlike existing multimodal approaches that primarily focus on 3D object detection, such as LiDAR-camera fusion in autonomous driving, SIM-Net bridges the gap between 2D image classification and 3D structural learning, enhancing object recognition by creating a more expressive feature space. Unlike previous models that rely on sensor-based 3D data, such as LiDAR or RGB-D cameras, our approach directly infers point clouds from 2D images. Additionally, while traditional CNNs primarily leverage color and texture cues, SIM-Net enhances classification by integrating shape-based reasoning, resulting in a feature-rich multimodal representation that improves generalization to unseen samples.

A prime application of our approach is the classification of digitized herbarium specimens. We conducted extensive experiments using diverse datasets comprising a wide range of plants. Our model demonstrated a significant improvement in classification accuracy compared to ResNet alone.

In summary, our main contributions in this paper are as follows:
- We introduce a pixel-to-point transformation pipeline that systematically converts segmented 2D objects into 3D point clouds.
- We propose a multimodal feature fusion network designed as a dual-branch architecture, with an image encoder and a point cloud encoder. These two complementary feature sets are fused into a unified latent space, enabling the model to learn both texture-based and shape-based characteristics simultaneously.
- We demonstrate that transforming 2D images into point clouds allows for the infusion of additional knowledge into the classification process.
- We show that point cloud-based architectures can surpass more complex traditional 2D classification models when applied to 2D image classification tasks.

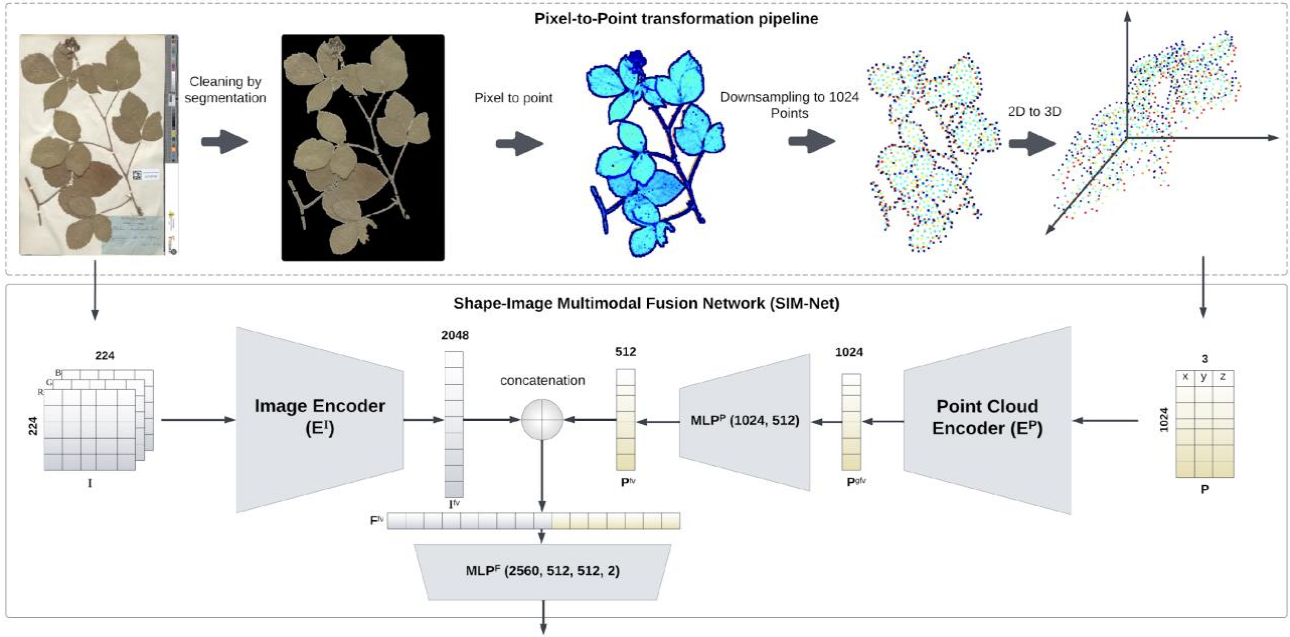

*Figure 3: 2D image to 3D point cloud data transformation pipeline and the Shape-Image Multimodal Fusion Network.*

The paper is organized as follows: Section 2 reviews related work. Section 3 details the methodology and processing techniques. Section 4 presents the experimental setup, data collections, findings and a discussion. Finally, Section 5 concludes with a summary of the results and potential directions for future research.

## 2. Related work

**2D image classification:**

2D image classification has advanced considerably over the last decade. AlexNet, [2] introduced key innovations such as the ReLU activation function, dropout layers to mitigate overfitting, and overlapping pooling. VGGNet [3], emphasized the depth, utilizing 16 convolutional layers. GoogLeNet/Inception [4] introduced the Inception module, which concatenates feature maps from filters of different sizes to capture multi-scale information efficiently. ResNet [5] introduced residual connections, allowing gradients to flow directly through networks exceeding 100 layers in depth. DenseNet [22] connects each layer to every other layer in a feed-forward fashion, maximizing information flow throughout the network. Transformer Models [6], originally developed for natural language processing, have been adapted for image classification through self-attention mechanisms that support parallelization and capture long-range dependencies. Vision Transformers [10] treat image patches as sequences for this purpose. Despite their general effectiveness, traditional 2D image classification techniques face substantial challenges when applied to digitized herbarium specimens, largely due to their sensitivity to background noise [18]. Figure 1 illustrates the wide variation in paper color, quality, and the presence of non-plant elements in herbarium images. Models such as ResNet and Vision Transformers display varying degrees of robustness to background noise and sensitivity to foreground-background contrast [18,19].

**3D point cloud classification**:

A point cloud is a collection of data points in space. Processing point clouds is challenging due their irregular structure, which requires networks that are invariant to point order, density variations, and spatial transformations. PointNet [7] addresses this challenge with a neural network specifically designed to process point clouds while maintaining permutation invariance, offering a streamlined and efficient approach. Despite its simplicity, PointNet achieves notable efficiency and effectiveness. PointNet++ [23] extends this architecture by applying PointNet recursively on a nested partitioning of the input point set, leveraging metric space distances to learn local features at progressively larger contextual scales. The Point Transformer [24] further enhances point cloud processing by extracting both local and global features through a novel local-global attention mechanism. PointMLP [25] adopts a pure residual MLP (Multi-Layer Perceptron) architecture without complex local geometric extractors yet delivers competitive performance. PointNeXt [26] builds upon PointNet++ by incorporating improved training strategies, an inverted residual bottleneck design, and separable MLPs, enabling efficient and scalable point cloud processing.

**2D-3D multimodal architectures**:

Numerous studies have explored strategies for fusing 2D images with 3D point clouds, employing a variety of approaches. However, most of them focus on improving neural network performance in 3D domains, such as 3D object detection [27,28] or 3D segmentation [29,30]. Poliyapram et al. [29] proposed the Point-wise LiDAR and Image Multimodal Fusion Network (PMNet) for 3D semantic segmentation of aerial point clouds by integrating aerial image features with LiDAR data. PMNet's architecture comprises two main components: a PointNet-based [7] backbone for point cloud feature extraction and a CNN Encoder-Decoder for image feature extraction. The point-wise fusion method combines features from both networks using a spatial correspondence table (XY coordinate index from the original point cloud data). This ensures that the extracted point-wise image features precisely mirror the point cloud structure in terms of the number of patches, points (N), and feature dimensions. The network merges 128 features from 2D aerial images and 128 features from the point cloud before feeding them into an MLP for final processing. Pradelle et al. [30] proposed a segmentation technique that leverages both 2D images and 3D point cloud data, eliminating the need for 3D ground truth and relying solely on 2D labels. This approach addresses the challenges of 3D ground truth labeling, which is labor-intensive and complex, in contrast to the simpler process of labeling 2D images. Their method employs a lightweight encoder to process features from a 3D point set, enhancing 2D segmentation results. The network accepts a point cloud, an image, and the corresponding projection matrix as inputs. It processes the point cloud with a 3D encoder to extract features for each visible point, which are then projected onto the image to create a 2D geometric feature map. This map is combined with the RGB image using a single-layer MLP and fed into a 2D segmentation network.

Yang et al. [34] introduced the Multimodal Interlaced Transformer (MIT) for weakly supervised 3D point cloud segmentation using only scene-level annotations. Their architecture employs two separate encoders: one for 2D multi-view images and another for 3D point clouds. The decoder performs interlaced 2D-3D cross-attention without requiring camera poses or depth maps. Notably, MIT alternates the roles of queries and key-value pairs across decoder layers to enable iterative fusion, enhancing each modality with features from the other. Zhao et al. [35] introduced SAFNet (Similarity-Aware Fusion Network), a late fusion framework for 3D semantic segmentation that adaptively fuses 2D image features and 3D point cloud data. Unlike traditional projection-based fusion methods, SAFNet addresses key challenges such as local mismatching and varying point density by computing geometric and contextual similarities between original and back-projected point clouds. These similarity metrics guide the fusion process, allowing the model to dynamically adjust the contribution of each modality per point.

Guo et al. [27] presented the Deep Multi-scale and Multi-modal Fusion (DMMF) for 3D object detection in autonomous driving, where cameras and LiDAR gather 3D spatial information. DMMF involves projecting the point cloud onto a Bird's Eye View (BEV) and extracting features from both the BEV map and the RGB image. These multi-modal features are then fused using a deep multi-scale fusion technique, followed by their application in a position regression and classification network for precise object classification and positioning. This method has shown impressive results on the KITTI dataset, especially in detecting cars and pedestrians, and has demonstrated significant accuracy improvements for more challenging data. Gu et al. [28] introduced

PointSee, a lightweight and flexible multimodal fusion solution to improve 3D object detection (3OD) networks. It Integrates LiDAR point clouds with scene images through a unique architecture comprising a hidden module (HM) that decorates LiDAR point clouds using 2D image information in an offline fusion manner. This leads to minimal or no adaptations of existing 3OD networks. A seen module (SM) enriches the LiDAR point clouds by acquiring point-wise representative semantic features, leading to enhanced performance of existing 3OD networks. This approach streamlines fusion without significant network modifications, offering a plug-and-play module for existing 3OD frameworks. Extensively validated on outdoor (KITTI) and indoor (SUN-RGBD) datasets, PointSee demonstrates marked improvements in performance, particularly when combined with 3DSSD. Its contributions lie in its novel modules (HM and SM), enhancing 3D object detection while maintaining network simplicity and efficiency. Similarly, SAMFusion [36] addresses 3D object detection in harsh environments (night, fog, snow) using RGB, NIR gated cameras, radar, and LiDAR. Its architecture incorporates depth-aware attention and a transformer operating in BEV space. SAMFusion targets the autonomous driving domain and relies entirely on physical sensors.

Chen et al. [37] tackle the task of depth completion from RGBD inputs by proposing a novel architecture that learns joint 2D-3D representations. Their core contribution is a modular 2D-3D fusion block composed of two domain-specific sub-networks: one operating in the 2D image space via traditional convolutions, and another in the 3D space using continuous convolutions on sparse point clouds. These branches extract appearance and geometric features independently, which are then fused in the image domain. By stacking these blocks, the network builds a hierarchical fusion of features across both modalities. In [31], the authors utilize 2D datasets to construct 3D facial data using a method that incorporates physical depth, which differs from our approach. Our objective is not to create a 3D representation of plants but to generate an alternate representation using point clouds. Distinctive aspects of our methodology include the representation of plant shapes in point clouds using pixel (x,y) coordinates and variations in color intensity, rather than physical depth. Furthermore, we address the unique challenges presented by herbarium datasets, as opposed to facial data, and employ specialized architectural and fusion techniques tailored to plant morphology. Additionally, works like [38] focus on hyperspectral image classification using combined 2D and 3D CNNs. These models operate on volumetric data (spatial + spectral), not on inferred point clouds, and are therefore not directly comparable to our pixel-to-point transformation pipeline.

In summary, the vast majority of recent 2D–3D fusion approaches are tailored toward 3D-centric tasks such as semantic segmentation or object detection, and they typically rely on explicit 3D input from physical sensors like LiDAR or RGB-D cameras. These methods are often designed to process entire scenes or multiple views and focus on leveraging depth or spatial priors grounded in real-world geometry. In contrast, our method introduces a novel perspective by performing 2D image classification using a synthetic 3D representation inferred directly from a single 2D image. Rather than leveraging real 3D sensor data, we transform segmented image regions into partial point clouds through a pixel-to-point conversion that encodes spatial coordinates and color information. This object-centric and sensor-free approach enables geometric reasoning without requiring depth supervision, multi-view consistency, or full-scene understanding. Our method remains distinct in both its objectives and input data modality, bridging a novel space between 2D vision and geometric reasoning for classification tasks.

## 3. Approach

Our goal is to improve image classification by fusing 2D images and 3D point cloud data in a multimodal architecture to identify the characteristics of objects of the same nature, such as herbarium images. The core of our approach involves transforming 2D images into 3D point clouds that represent the targeted objects. To achieve this, we first use segmentation techniques to extract masks, which we transform into 3D object shape point clouds.

The original 2D images and their corresponding 3D object shape point clouds form pairs of (image, object shape point cloud) that are integrated into a multimodal fusion architecture (SIM-Net) comprising two encoders: an image encoder (ResNet) and a point cloud encoder (PointNet). The 3D shape representations provide crucial geometric and spatial information not readily discernible in 2D images, thereby enriching the feature space accessible to the learning model. For herbarium images, segmentation enables the extraction of the plant from the herbarium sheets while removing background elements.

### 3.1. Image to point cloud data transformation

The transformation pipeline for converting herbarium sheet images into 3D point clouds of 1024 points, as illustrated in Figure 3, begins with a segmentation phase. The original herbarium image is processed using the U-Net [32] network, selected for its proven ability to extract detailed features from complex backgrounds. The U-Net training dataset was prepared using a color segmentation approach, detailed in four primary stages as described in Ariouat et al. [33], and employed YOLOv7-ag [17] for detecting non-plant elements. After reducing noise by detecting and removing non-plant elements, generating masks via color segmentation, and applying morphological refinements, the final masks are used to extract plant pixels from the original images. This systematic process has resulted in a curated dataset of 2225 images and their corresponding masks, which was used to train U-Net.

The U-Net model segments each herbarium image into a binary mask, precisely isolating the plant material from the background. This mask is then projected onto the original image to selectively extract plant pixels while preserving the morphology of the specimen. Non-plant pixels are set to black by assigning their red, green and blue $(R, G, B)$ color values to $(R, G, B) = (0,0,0)$.

An image $I \in R^{H \times W \times C}$ is a tensor, where two dimensions (the height $H$ and the width $W$) represent the spatial coordinates $(x, y)$ of a pixel, and the third dimension (the number of color channels $C$ represents its color $(R, G, B)$. Each pixel (tensor element) $I(x, y) = (R_{xy}, G_{xy}, B_{xy})$ is a triplet of color intensities $(R_{xy}, G_{xy}, B_{xy})$, typically ranging from 0 to 255. For each pixel $I(x, y)$ with $(R_{xy}, G_{xy}, B_{xy}) \neq (0,0,0)$, the transformation to a 3D point starts by a mapping function that maps the pixel to a new point $p(x_p, y_p)$, where $x_p = x$ and $y_p = y$, in a two-dimensional space, using the spatial coordinates.

This transformation generates a large point cloud $P$, where $|P| < H \times W$, in a two-dimensional space from the image pixels, using the spatial coordinates of each eligible pixel. The resulting point cloud is depicted in blue (*cf.* Figure 3). During the downsampling stage, the point cloud is pruned to retain only 1024 points. Given a point cloud $P = \{p_1, p_2, \ldots, p_n\}$, where $|P| = n$, we use a Farthest Point Sampling (FPS) algorithm for downsampling $P$ to a smaller subset $P^m = \{p'_1, p'_2, \ldots, p'_m\}$, where $m = 1024$, that best represents the original set $P$ in terms of spatial distribution. Each new point is chosen based on its maximum distance from the previously selected points $\{p'_1, p'_2, \ldots, p'_{j-1}\}$, ensuring that each new addition is the farthest from the existing subset with respect to the remaining points. This method of sampling distinctly contrasts with random sampling, as FPS achieves a more comprehensive coverage of the entire set of points, maintaining effectiveness even with a limited number of centroids. Furthermore, the FPS algorithm adapts to the data distribution, ensuring that the reduced subset retains the essential features of the original point cloud. This downsampling is pivotal to reduce computational complexity and to focus on significant morphological features, facilitating the deployment of deep learning models like PointNet. The final stage involves adding the third dimension $z_p$ to each point $p(x_p, y_p)$ in the point cloud. This is achieved by translating the color information of each corresponding eligible pixel $I(x, y)$ with $(R_{xy}, G_{xy}, B_{xy}) \neq (0,0,0)$ into a spatial dimension $z_p$. Specifically, $z_p$ is assigned as the average of the pixel's color values:

$$p(x_p, y_p, z_p) \text{ where } x_p = x_p, y_p = y_p \text{ and } z_p = \frac{R_{xy} + G_{xy} + B_{xy}}{3}$$

From a conceptual standpoint, the rationale for transforming 2D image pixels into 3D point cloud representations lies in the objective of capturing the global shape structure of the target object rather than its real appearance alone. While convolutional neural networks are highly effective at extracting localized texture-based features from 2D images, they remain inherently sensitive to background clutter, lighting variations, and pose. In contrast, architectures like PointNet are specifically designed to learn from unordered point sets and are good at modeling the overall geometric configuration of an object—its shape— independently of color or background context. In our case, the inferred point cloud is not intended to represent true physical depth, but rather to encode shape cues derived from the spatial distribution and intensity of pixels within the targeted object (the plant).

### 3.2. Image encoder

Image encoders play a critical role in multimodal architectures due to their ability to process and interpret complex data types. Encoders like ResNet [5] are particularly effective for processing 2D images and extracting features such as edges and textures. In our approach, we employ the ResNet101 encoder $E^I$ (see Figure 3) within the SIM-Net architecture to extract detailed point-wise image features from 224 x 224 images. These features are encoded into an image feature vector $I^{fv}$, where $|I^{fv}| = 2048$, as illustrated in Figure 3.

### 3.3. Point cloud encoder

Point clouds are sets of points in 3D space. Each point is defined by its x, y, and z coordinates and may include additional attributes such as color or intensity. Networks specifically designed for 3D data, such as PointNet [7], are required to process the derived object shape point clouds. In SIM-Net, we leverage the classification component of the PointNet network as a point cloud encoder $E^P$. This encoder takes 1024 points as input, processes each point individually through input and feature transformations and fully connected layers and, aggregates point features by using max pooling to produce an order-invariant global feature vector $P^{gfv}$, where $|P^{gfv}| = 1024$ :

$$P^{gfv}(p_1, p_2, \ldots, p_m) = g(h(p_1), h(p_2), \ldots, h(p_m))$$

where, $m = 1024$ is the number of points in $P$, $h$ is an MLP function, $g$ is a global max-pooling function.

The difference between 2D images and point clouds in terms of the features captured by neural networks like ResNet [5] and PointNet [7] is quite significant due to their inherent nature. Unlike 2D images, point clouds capture the spatial structure and shape of objects in three-dimensional space, providing information about geometry and volume. PointNet is capable of recognizing patterns in the spatial arrangement of points. It can extract insights about the shape and structure of objects, aspects that 2D images alone cannot provide.

### 3.4. Fusing RGB features with 3D global features

Fusion strategies are central to the effectiveness of multimodal architectures, as they enable the integration of complementary information from different data modalities. These strategies typically combine the outputs of two specialized neural networks to achieve a unified prediction task [30,29,27,28]. In SIM-Net, the fusion function $F$ is implemented by concatenating the image feature vector $I_i^{fv} = E^I(i)$ of an image $i$, where $E^I$ is the image encoder function, and the point cloud feature vector $P_i^{fv} = MLP^p(P_i^{gfv})$ of a point cloud $P_i = T(i)$ derived from the image $i$, where $MLP^P$ is an MLP function, $P_i^{gfv} = E^P(T(i))$ is the point cloud global feature vector, $E^P$ is the point cloud encoder function and $T(i)$ is the 2D-to-3D image transformation function.

$$F_i^{fv} = F(I_i^{fv} \oplus P_i^{fv})$$

$F_i^{fv}$ represents the fusion feature vector with a size of 2056. This vector is fed into the $MLP^F$ MLP function for classification. $MLP^F$ is an MLP that contains three fully connected layers with ReLU activation function, batch normalization and dropout function. Finally, a sigmoid activation function for binary classification was carried out as illustrated in the SIM-Net architecture (Figure 3). The combination of ResNet global features with PointNet features allows the model to exploit both detailed surface information and 3D spatial relationships. This fusion creates a more robust and comprehensive feature representation, enabling the model to better classify complex structures that require the integration of surface details and geometric context.

### 4. Experiments

In this section, we present a set of experiments designed to evaluate the effectiveness of the PointNet, PointNet++ and SIM-Net models using the point clouds derived from herbarium 2D images, with ResNet serving as the baseline for comparison. The first set of experiments (*cf.* Table 2 and Table 3) assesses the performance of PointNet [7] and PointNet++ [23] relative to ResNet101 [5] trained from scratch. PointNet++

extends the original PointNet through a hierarchical architecture that applies its core principles recursively across nested partitions of the input point set[2].

We further evaluate how point quality impacts model performance by conducting tests on both a curated set of high-quality point clouds and a broader dataset that may include lower-quality point clouds. The quality of the point clouds depends directly on the image segmentation. Additionally, we explore the benefits of enriching point clouds by appending the three RGB color values to each point, thereby increasing the dimensionality of the point cloud representation from three to six.

We then present a series of experiments evaluating the performance of point clouds within the SIM-Net multimodal architecture, which uses ResNet101 as the image encoder and PointNet as the point cloud encoder (*cf.* Table 4 and Table 5). We benchmark SIM-Net against ResNet101 pretrained on the ImageNet-1k dataset. In this second series of experiments, we examine how segmentation quality–and consequently, point cloud quality– influences SIM-Net performance. We compare model results across datasets containing various plant traits, using both well-segmented samples and datasets that may include lower-quality segmentations.

### 4.1. Experiments settings

In this study, we evaluated four distinct models: ResNet101, PointNet, PointNet++, and SIM-Net. The experiments were conducted using four Nvidia A100 GPUs. For the ResNet101 and SIM-Net models, we employed a range of batch sizes [8, 16, 32, 64, 128] across all training sessions. PointNet and PointNet++ were trained using a fixed batch size of 24. All models were trained for 100 epochs, employing the Adam optimizer with an initial learning rate of 0.001 and a weight decay set at 1e-4. We applied a StepLR scheduler to systematically reduce the learning rate by 30% every 20 epochs. The cross-entropy loss function was used for network optimization.

To mitigate overfitting in the PointNet and PointNet++ models, we applied several data augmentation techniques, including random rotations around the z-axis and the addition of random noise to the point cloud data. For the ResNet model, we used three augmentation strategies, namely Random Horizontal Flip, Random Vertical Flip, and Random Rotation set at 90 degrees. These techniques were chosen to increase variability and complexity within the training datasets.

Our evaluation metrics were focused on accuracy and F1-score to ensure a thorough assessment of each model's performance. All implementations were carried out using the Pytorch framework.

### 4.2. Datasets

In our study, we used an initial set of 4,005 images annotated for the presence or absence of five distinct plant traits: thorns (sharp structures on branches or stems), presence of fruits, leaves with acuminate tips, infructescence (arrangement of elementary fruits, like grapes), and leaves with an acute base. These images were processed through a segmentation pipeline, resulting in a second dataset of 4,005 segmented images. This segmented dataset was subsequently used for 2D-to-3D transformation, producing a corresponding dataset of 4,005 point-clouds. However, we faced two major challenges: significant class imbalance in trait annotations and inconsistent segmentation quality across some images. To address class imbalance, we created a *Complete Datasets* series, consisting of two balanced datasets per trait: one containing segmented images and one containing unsegmented images. For each image dataset, we also generated a corresponding point cloud dataset. To study the impact of segmentation quality, we selected well-segmented images from the Complete Datasets and generated their corresponding point clouds, forming the *Selected Datasets* series. Figure 4 provides a visual comparison of image segmentation quality and its impact on point cloud generation. The top row shows an example of high-quality segmentation resulting in a well-formed point cloud, which represents the standard we aim to maintain in our Selected Datasets series.

---

[2] PointNet falls short in recognizing local structures within the metric space in which the points reside, which restricts its capacity to discern intricate patterns and adapt to multifaceted scenes. By leveraging metric space distances, PointNet++ is adept at capturing and learning local features at progressively larger contextual scales.

Table 1: Datasets distribution for complete and selected datasets.

| Trait | Complete Dataset | | Selected Dataset | |
|---|---|---|---|---|
| | Train | Val | Train | Val |
| Thorns | 2426 | 500 | 1786 | 380 |
| Fruits | 2776 | 580 | 1812 | 380 |
| Leaves with acuminate tips | 2360 | 500 | 1648 | 340 |
| Infructescence | 2340 | 480 | 1570 | 340 |
| Leaves with an acute base | 2308 | 500 | 1360 | 280 |

Table 1 presents the size of training and validation sets of the Complete and Selected datasets series. In contrast, the bottom row illustrates an example of lower-quality segmentation, which leads to a correspondingly lower-quality point cloud transformation. Such cases are representative of the broader range of image qualities found in the Complete Datasets collection.

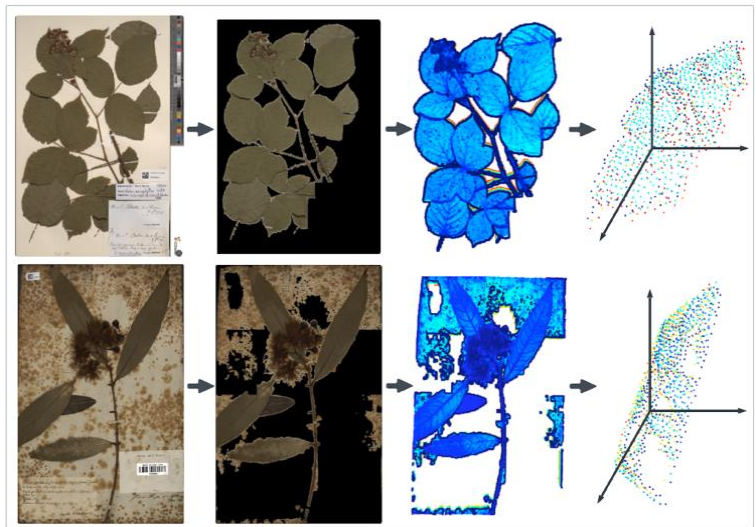

Figure 4: Transformation examples showing accurate point cloud generation (top) and less precise point cloud generation (bottom).

We created a collection of 40 unimodal datasets, each comprising image data and point cloud data. These datasets are derived from two distinct series: Complete and Selected. For each of the five traits studied, we generated eight distinct datasets —four from the Complete series and four from the Selected series— consisting of unsegmented images, segmented images, point clouds generated from unsegmented images, and point clouds generated from segmented images. Each series thus contributes 20 unique datasets, yielding a total of 40 unimodal datasets. In addition, we generated 20 multimodal datasets, resulting in an additional 20 datasets. The composition and naming conventions for each multimodal dataset across the five traits are described below:

1. **Complete Image Point cloud Dataset (CIPD)**: Combines the full set of original unsegmented images with the point clouds generated from their corresponding segmented images.
2. **Complete Segmented Image Point cloud Dataset (CSIPD)**: Pairs the entire collection of segmented images with the point clouds derived from the same segmented images.
3. **Selected Segmented Image Point cloud Dataset (SSIPD)**: Merges the selected segmented images with the point clouds generated from these accurately segmented images.
4. **Selected Image Point cloud Dataset (SIPD)**: Consists of the selected set of original unsegmented images that correspond to the well-segmented images, along with their associated point clouds from the well-segmented counterparts.

Additionally, we generated two enriched point cloud datasets for each trait–one within the selected series and one within the complete series. These datasets consist of six-dimensional point clouds in which the RGB color values of each corresponding pixel are appended to the original $(x, y, z)$ coordinates of each point. This integration transforms each pixel into one point with $(x, y, z, r, g, b)$ coordinates. We use these enriched datasets to investigate the effects of augmenting the original point clouds with additional information.

Altogether, this expansion yields a total of 70 distinct datasets used in our experiments, providing a rich foundation for comprehensive analysis.

*Table 2: Accuracy (from scratch) using unsegmented selected images and selected point cloud datasets.*

| Trait | Images | Point cloud with 3 dimensions (x,y,z) | | | | Point cloud with 6 dimensions (x,y,z,r,g,b) | | | |
|---|---|---|---|---|---|---|---|---|---|
| | ResNet101 | PointNet | Δ | PointNet++ | Δ | PointNet | Δ | PointNet++ | Δ |
| Thorns | 75.46 | 61.03 | -14.43 | 83.05 | +7.59 | 85.73 | +10.27 | 92.90 | +17.44 |
| Fruits | 54.09 | 54.01 | -0.08 | 59.34 | +5.25 | 54.71 | +0.62 | 58.58 | +4.49 |
| Leaves with acuminate tips | 58.82 | 59.72 | +0.9 | 68.61 | +9.79 | 64.72 | +5.9 | 73.61 | +14.79 |
| Infructescence | 54.87 | 56.11 | +1.24 | 63.33 | +8.46 | 57.22 | +2.35 | 59.72 | +4.85 |
| Leaves with an acute base | 59.86 | 56.52 | -3.34 | 65.83 | +5.97 | 67.29 | +7.43 | 71.45 | +11.59 |

*Table 3: Accuracy (from scratch) using unsegmented Complete images and complete point cloud datasets.*

| Trait | Images | Point cloud with 3 dimensions (x,y,z) | | | | Point cloud with 6 dimensions (x,y,z,r,g,b) | | | |
|---|---|---|---|---|---|---|---|---|---|
| | ResNet101 | PointNet | Δ | PointNet++ | Δ | PointNet | Δ | PointNet++ | Δ |
| Thorns | 72.00 | 59.08 | -12.92 | 86.23 | +14.23 | 84.36 | +12.36 | 93.76 | +21.76 |
| Fruits | 53.62 | 53.50 | -0.12 | 56.50 | +2.88 | 58.33 | +4.71 | 59.50 | +5.88 |
| Leaves with acuminate tips | 63.00 | 57.89 | -5.11 | 75.11 | +12.11 | 68.21 | +5.21 | 73.96 | +10.96 |
| Infructescence | 55.63 | 52.04 | -3.59 | 62.03 | +6.4 | 56.25 | +0.62 | 68.87 | +13.24 |
| Leaves with an acute base | 59.20 | 55.20 | -4.00 | 66.34 | +7.14 | 61.30 | +2.10 | 68.49 | +9.29 |

*Table 4: Evaluation metrics on unsegmented images with point cloud datasets.*

| Trait | Selected Image Point cloud (SIPD) | | | | | | Complete Image Point cloud (CIPD) | | | | | |
|---|---|---|---|---|---|---|---|---|---|---|---|---|
| | SIM-Net | | ResNet101 | | Δ | | ResNet101 | | SIM-Net | | Δ | |
| | Acc | F1 | Acc | F1 | Acc | F1 | Acc | F1 | Acc | F1 | Acc | F1 |
| Thorns | 95.75 | 95.74 | 95.53 | 94.02 | +0.22 | +1.72 | 95.00 | 95.01 | 95.20 | 93.47 | -0.20 | +1.54 |
| Fruits | 66.56 | 69.65 | 61.58 | 66.08 | +4.98 | +3.57 | 64.34 | 69.52 | 64.66 | 67.22 | -0.32 | +2.3 |
| Leaves with acuminate tips | 79.84 | 79.19 | 77.94 | 73.02 | +1.9 | +6.17 | 81.40 | 81.44 | 77.00 | 73.92 | +4.40 | +7.52 |
| Infructescence | 70.19 | 71.35 | 60.29 | 66.67 | +9.9 | +4.68 | 73.05 | 74.24 | 69.37 | 69.10 | +3.68 | +5.14 |
| Leaves with an acute base | 74.67 | 75.84 | 70.71 | 71.61 | +3.96 | +4.23 | 71.80 | 73.46 | 71.40 | 72.66 | +0.40 | +0.80 |

*Table 5: Evaluation metrics on segmented images with point cloud datasets.*

| Trait | Selected Segmented Image Point cloud (SSIPD) | | | | | | Complete Segmented Image Point cloud (CSIPD) | | | | | |
|---|---|---|---|---|---|---|---|---|---|---|---|---|
| | SIM-Net | | ResNet101 | | Δ | | SIM-Net | | ResNet101 | | Δ | |
| | Acc | F1 | Acc | F1 | Acc | F1 | Acc | F1 | Acc | F1 | Acc | F1 |
| Thorns | 92.61 | 92.55 | 93.4 | 91.17 | -0.79 | +1.38 | 95.9 | 95.79 | 94.2 | 92.86 | +1.7 | +2.93 |
| Fruits | 64.88 | 67.86 | 60.95 | 62.33 | +3.93 | +5.53 | 64.84 | 67.03 | 60.17 | 67.28 | +4.67 | -0.25 |
| Leaves with acuminate tips | 81.82 | 81.4 | 70.59 | 70.62 | +11.23 | +10.78 | 82.27 | 82.33 | 79.4 | 78.88 | +2.87 | +3.45 |
| Infructescence | 66.35 | 69.51 | 63.72 | 57.23 | +2.63 | +12.28 | 70.2 | 72.32 | 70.83 | 72.87 | -0.63 | -0.55 |
| Leaves with an acute base | 73.87 | 74.92 | 71.68 | 74.43 | +2.19 | +0.49 | 71.84 | 73.56 | 70.0 | 69.42 | -1.16 | +4.14 |

### 4.3. Discussion

The experimental results are presented in Tables 2, 3, 4, 5 and 11. Tables 2 and 3 report the accuracy achieved by each model (ResNet101, PointNet, and PointNet++), with the delta (Δ) columns indicating the accuracy difference between the point cloud models (PointNet and PointNet++) and ResNet101. The tables 4 and 5, in contrast, present the highest accuracy (Acc) and F1-score (F1) values achieved by both SIM-Net and ResNet101. It is important to note that the Acc and F1-score values listed for a given model on the same row may originate from different training runs, as we selected the highest accuracy, and the highest F1-score obtained across three training iterations for each model on each dataset. The delta columns in these tables indicate the performance gain or loss of SIM-Net relative to ResNet101.

An initial observation from Tables 2 and 3 concerns the accuracy of ResNet101, PointNet, and PointNet++ models trained from scratch. We observe that PointNet consistently underperforms relative to ResNet101 across both the Selected and Complete dataset series. However, PointNet++ achieves up to a 14% performance improvement over ResNet101 for the Thorns trait (see Table 3). This suggests that with large 3D point cloud datasets derived from 2D images– comparable in scale to ImageNet-1K– models such as PointNet and PointNet++ can surpass traditional models like ResNet.

A key finding from this first series of experiments is the performance boost obtained by enriching point clouds with additional information—specifically, by appending RGB color values to the (x, y, z) coordinates. This enhancement results in a 17% performance gain for PointNet++ over ResNet101 for the Thorns trait in the Selected datasets (Table 2), and an even larger 21% improvement in the Complete datasets (Table 3). These results highlight the potential of embedding additional image-based information into the point cloud representation.

In the second series of experiments (Tables 4 and 5), we observe that the multimodal architecture (SIM-Net) consistently achieves higher accuracy in 80% of the test cases (16 out of 20) and higher F1-score in 90% of the cases (18 out of 20). The first major insight is that SIM-Net outperforms ResNet101 pretrained on ImageNet-1K in nearly all scenarios. This holds true for both segmented and unsegmented images, across both the Selected and Complete datasets, with only 5 exceptions out of 40 cases. It is important to note that ResNet101 serves as a unimodal baseline that relies solely on 2D image data, whereas SIM-Net integrates both 2D images and their corresponding inferred 3D point clouds. This comparison demonstrates the added value of incorporating 3D geometric features into the classification process.

A second key observation is that SIM-Net performs better on the Selected datasets series than on the Complete series, particularly for unsegmented images (see Table 4). This improvement can be attributed to the superior quality of the point clouds in the Selected datasets compared to those in the Complete series. This trend remains consistent even for segmented images (see Table 5), reinforcing the conclusion that point cloud quality— driven by segmentation accuracy — plays a critical role in enhancing the model's performance.

From a theoretical perspective, the performance improvement observed with the 2D-to-3D transformation can be attributed to the complementarity between 2D visual features and 3D geometric representations. Traditional 2D image architectures primarily capture local texture, color, and edge-based features within a fixed receptive field. These features are highly sensitive to background context, lighting conditions, and specimen placement, which can vary significantly across images. In contrast, transforming segmented 2D images into 3D point clouds produces a spatial representation of object shape that abstracts away background variation while preserving morphological consistency. This structural representation encodes topological and geometric relationships between object parts, which are invariant to appearance-level noise. In a multimodal setting, the 2D and 3D branches provide orthogonal yet complementary feature spaces: the image encoder captures discriminative texture-based cues, while the point cloud encoder encodes global and local shape priors. This fusion enhances the model's capacity to distinguish fine-grained inter-class differences and generalize across varying background conditions–especially when classification relies on subtle morphological traits.

Despite SIM-Net's promising results, several challenges present opportunities for improvement. Segmentation quality remains crucial, as errors in segmentation directly affect point cloud generation and, consequently, classification accuracy. Enhancing segmentation models through self-supervised learning or attention mechanisms could improve robustness. Additionally, multimodal fusion increases computational complexity; optimizing architectures with lightweight feature extractors or more efficient fusion strategies could help reduce resource demands. Another open challenge is generalization beyond herbarium datasets. Domain adaptation techniques may be required to enhance performance across diverse object categories. Scalability is another concern, as the 2D-to-3D transformation process introduces computational overhead. Applying efficient downsampling techniques or hybrid processing strategies could enhance efficiency for large-scale datasets. Furthermore, while point clouds encode spatial structure, they lack true depth information. Exploring monocular depth estimation or benchmarking SIM-Net against alternative multimodal fusion techniques could yield valuable insights. In summary, improvements in segmentation, computational efficiency, generalization, and depth representation could further enhance SIM-Net's scalability and applicability, making it a more versatile classification solution.

## 5. Conclusion

The foundation of our approach lies in transforming the same dataset into multiple representations to leverage the strengths of different architectures for feature extraction. In this study, we convert 2D herbarium images into point clouds, enabling the use of a multimodal architecture that combines a convolutional neural network for detailed pixel-level analysis with a point cloud-based neural network for interpreting geometric structures. This methodology significantly improves the classification performance of herbarium images, particularly in complex and nuanced scenarios where traditional 2D methods might struggle.

A crucial aspect of our approach is its dependence on the quality of both the segmentation and the resulting point clouds. Higher segmentation quality consistently leads to improved performance. This relationship demonstrates the importance of precise and accurate initial data processing within our pipeline.

Looking ahead, several avenues could further enhance the performance of our model. One promising direction is the integration of advanced point cloud encoders such as PointNeXt, PointMLP, and Point Transformer. These models are particularly effective in processing point cloud data. Additionally, optimizing the data transformation pipeline presents a significant opportunity for improvement. By incorporating additional knowledge into the point clouds, we can enrich the data and potentially achieve more nuanced and accurate classification results. This could involve advanced techniques in data augmentation techniques or the integration of supplementary features to help the model better capture complex patterns.

From a theoretical perspective, it would be intriguing to understand why the addition of 3D modality enables the creation of more efficient representations, given that the initial information is entirely 2D and no new external information is added–apart from the explicit inclusion of pixel position coordinates. This raises interesting questions about how the 2D-to-3D transformation enhances the informational richness of the representation.

## Acknowledgements

This work was part of tasks within the e-Col+ project (ANR-21-ESRE-0053). This project was provided with computing HPC and storage resources by GENCI at IDRIS thanks to the grant 2023-A0150114385 on the supercomputer Jean Zay's A100 partition.

## Conflict of interest statement
The authors declare that they have no known competing financial interests or personal relationships that could have appeared to influence the work reported in this paper.

## Funding information
Agence Nationale de la Recherche (ANR), e-Col+ Project, Grant/Award Number: ANR-21-ESRE-0053; GENCI, HPC and Storage Resources, Grant/Award Number: 2023-A0150114385

## Data availability statement
The data that support the findings of this study are available from the corresponding author upon request and will be released publicly upon acceptance.

## Code availability statement

The source code used in this study is available from the corresponding author upon request and will be released publicly upon acceptance.

## Author Contributions

Youcef Sklab led the conceptualization, data curation, formal analysis, methodology development, and software implementation. He also managed the project, funding acquisition, provided resources, supervised the work, and contributed to validation, visualization, and both the original draft and subsequent revisions. Hanane Ariouat contributed to data curation, resource management, software development, and writing of the original draft. Eric Chenin was involved in data curation, funding acquisition, and project administration. Edi Prifti contributed to funding acquisition, provided resources, supervised the work, and participated in manuscript revision. Jean-Daniel Zucker contributed to the funding acquisition, supervised the research, and supported the review and editing of the manuscript.

## Appendix

In Appendix, we provide additional information regarding four additional tests we conducted to evaluate the effectiveness of incorporating $x, y, z$ coordinates into point clouds in our approach.

### A. Extensive Evaluation with State-of-the-Art Models

The table Table 6 presents a comparative evaluation of eight state-of-the-art deep learning models (BEiT [39], CoAtNet [40], ConvNeXt [41], ConvNeXtv2 [42], ConViT [43, ] ViT [6], Swin [44] and DaViT [45]), assessed in two experimental settings: (1) single-modality classification using only 2D images and (2) multimodal fusion, where each model was integrated into our architecture (SIM-Net) as an image encoder in place of the originally used ResNet. The evaluation covers the same five botanical traits: Leaves with acuminate tips, Leaves with an acute base, Infructescence, Fruits, and Thorns. For each model and each setting, we conducted three independent runs from scratch, and the values presented in the table correspond to the best accuracy obtained across these runs. We run the experiments on an Nvidia A100 card, with a batch size set to 128. Each model underwent training for 100 epochs, employing the Adam optimizer with an initial learning rate of 0.001 and a weight decay set at 1e-4. A StepLR scheduler was applied, systematically reducing the learning rate by 30% every 20 epochs. The cross-entropy loss function was used for network optimization. The Delta column in the table indicates the performance gain (or loss) achieved by integrating the models into our multimodal fusion network (SIM-Net) compared to its performance in a single-modality setting.

Table 6: Comparative performance of eight state-of-the-art deep learning models in single-modality (2D images) and multimodal (2D + 3D fusion) settings for botanical trait classification.

| Model | Leaves with acuminate tips | | | Leaves with an acute base | | | Infructescence | | | Fruits | | | Thorns | | |
|---|---|---|---|---|---|---|---|---|---|---|---|---|---|---|---|
| | Single | SIM-Net | Δ | Single | SIM-Net | Δ | Single | SIM-Net | Δ | Single | SIM-Net | Δ | Single | SIM-Net | Δ |
| BEiT | 63.59 | 64.12 | +00.53 | 61.29 | 62.37 | +01.08 | 54.57 | 55.75 | +01.18 | 65.62 | 57.52 | -08.10 | 73.57 | 73.46 | -00.11 |
| CoAtNet | 50.59 | 58.53 | +07.94 | 50.54 | 59.50 | +08.96 | 50.15 | 54.18 | +04.03 | 50.92 | 57.78 | +06.86 | 59.89 | 75.20 | +15.31 |
| ConvNeXt | 57.65 | 61.76 | +04.11 | 50.54 | 62.37 | +11.83 | 53.10 | 54.28 | +01.18 | 70.71 | 57.78 | -12.93 | 67.01 | 73.09 | +06.08 |
| ConvNeXt v2 | 52.23 | 59.41 | +07.18 | 50.18 | 60.93 | +10.75 | 51.34 | 54.28 | +02.94 | 51.72 | 56.73 | +05.01 | 56.99 | 68.60 | +11.61 |
| ConViT | 55.59 | 59.71 | +04.12 | 58.78 | 60.02 | +01.24 | 53.98 | 54.87 | +00.89 | 67.81 | 57.52 | -10.29 | 69.39 | 68.34 | -01.05 |
| ViT | 59.71 | 61.18 | +01.47 | 60.44 | 61.29 | +00.85 | 55.16 | 56.64 | +01.48 | 70.46 | 58.05 | -12.41 | 74.41 | 65.80 | -08.61 |
| Swin | 53.24 | 60.00 | +06.76 | 50.18 | 63.80 | +13.62 | 52.51 | 54.87 | +02.36 | 52.31 | 57.52 | +05.21 | 68.07 | 61.48 | -06.59 |
| DaViT | 51.42 | 56.76 | +05.34 | 50.54 | 60.93 | +10.39 | 51.15 | 53.69 | +02.54 | 51.13 | 56.20 | +05.07 | 58.35 | 72.30 | +13.95 |

Our analysis demonstrates that SIM-Net consistently improves classification accuracy across a wide range of visual traits, particularly for Leaves with acuminate tips, Leaves with an acute base, and Infructescence. In most cases, SIM-Net outperforms the single-modality (2D-only) baselines. The most notable improvements are observed with CoAtNet, ConvNeXtV2, and Swin, for all the traits with accuracy gains reaching up to +13.62% for Leaves with an acute base in the Swin model. For the Thorns and Fruits traits, SIM-Net achieves better performance in 4 out of 8 models. Significant gains are recorded for CoAtNet and DaViT on Thorns, with improvements of +15.31% and +13.95%, respectively. However, not all results are positive. A slight performance drop is observed for BEiT on Thorns (–0.11%), and more notably for ViT (–8.61%). These decreases may be attributed to segmentation errors during the point cloud generation process, which can affect model input quality. In the Fruits category, SIM-Net also shows mixed outcomes, with some models such as ConvNeXt and ViT experiencing substantial drops of –12.93% and –12.41%, respectively. This drop in performance may stem from the fact that fruits are often light-colored and visually like the background (e.g., paper), leading to segmentation errors. In such cases, portions of the fruit may be misclassified as background, resulting in incomplete or noisy point clouds and degraded classification accuracy. Despite these challenges, SIM-Net consistently provides substantial performance gains across several traits and architectures. The structural traits, in particular, benefit the most from multimodal fusion. Models such as ConvNeXt (on Leaves with an acute base, +11.83%) and CoAtNet (on Leaves with acuminate tips, +7.94%) showcase the added value of incorporating 3D geometric cues alongside 2D visual features.

To further investigate the benefits of more advanced multimodal fusion mechanisms, we extended the SIM-Net architecture by integrating a cross-attention mechanism. In the adapted architecture, both point cloud and image features are first projected into a 512-dimensional embedding space through linear transformations and reshaped as single-token sequences. We explored three cross-attention configurations: i) in the first variant (SIM-$Net_{CA}$, Table 7), the point cloud features act as the query, and the image features serve as key and value; ii) in the second variant (SIM-$Net_{CA}$, Table 8), the image features act as the query, and the point cloud features serve as key and value; iii) in the third variant (SIM-$Net_{BCA}$, Table 9), we implemented a bidirectional cross-attention mechanism, where both directions of attention (point cloud to image and image to point cloud) are computed independently, and their outputs are concatenated before fusion. The attention outputs in each case are passed through a fully connected layer for classification. Batch normalization and dropout are also applied for regularization. We evaluated these cross-attention variants under the same experimental conditions as SIM-Net and ResNet101, using two dataset configurations: SIPD and CIPD.

The results are reported in Tables 7, 8, and 9. Overall, the use of cross-attention introduces variable effects across traits and datasets. While the single-direction cross-attention variants (Tables 7 and 8) occasionally lead to improvements on specific traits such as Thorns, they tend to degrade performance on several others, particularly for Fruits and Leaves with acuminate tips. Among these two single-direction variants, the configuration where the image acts as the query generally performs slightly better than the reverse. The bidirectional cross-attention variant (Table 9), which fuses the outputs of both attention directions, also shows a mixed impact: it achieves improvements on some traits and datasets (notably Thorns on CIPD) but again does not consistently outperform the baseline SIM-Net across all traits. These observations suggest that while cross-attention can enhance multimodal fusion in certain cases, its effectiveness is highly trait- and dataset-dependent and further tuning or more adaptive attention mechanisms may be required to obtain robust gains. Moreover, as in earlier experiments, the relatively small dataset size may limit the full potential of more sophisticated attention-based mechanisms.

More broadly, our results confirm that the integration of point cloud data consistently improves classification accuracy for traits with complex geometric structures. Notably, ResNet101 remains a strong image encoder across all settings. As shown in Table 6, SIM-Net equipped with ResNet101 outperforms alternative variants using other backbones, reinforcing the idea that deep convolutional networks offer excellent generalization in low-data regimes—even compared to more recent attention-based approaches—when combined with complementary 3D representations.

Table 7: Comparison of SIM-Net and cross-attention variants on the Selected Image Point Cloud Dataset (SIPD).

| Trait | SIM-Net | | Point Cloud as Query – Image as Key/Value | | | | Image as Query – Point Cloud as Key/Value | | | |
|---|---|---|---|---|---|---|---|---|---|---|
| | | | SIM-$Net_{CA}$ | | Δ | | SIM-$Net_{CA}$ | | Δ | |
| | Acc | F1 | Acc | F1 | Acc | F1 | Acc | F1 | Acc | F1 |
| *Thorns* | 92.61 | 92.55 | 94.46 | 94.43 | +1.85 | +1.88 | 94.46 | 94.52 | +1.85 | +1.97 |
| *Fruits* | 64.88 | 67.86 | 60.69 | 66.78 | -4.19 | -1.08 | 60.16 | 66.78 | -4.46 | -1.08 |
| *Leaves with acuminate tips* | 81.82 | 81.4 | 76.76 | 75.08 | -5.06 | -6.32 | 72.35 | 74.26 | -9.47 | -7.14 |
| *Infructescence* | 66.35 | 69.51 | 61.65 | 66.67 | -4.7 | -2.84 | 64.03 | 67.5 | -2.32 | -2.01 |
| *Leaves with an acute base* | 73.87 | 74.92 | 69.53 | 74.32 | -4.34 | -0.6 | 71.33 | 72.4 | -2.54 | -2.52 |

Table 8: Comparison of SIM-Net and cross-attention variants on the Complete Image Point Cloud Dataset (CIPD).

| Trait | SIM-Net | | Point Cloud as Query – Image as Key/Value | | | | Image as Query – Point Cloud as Key/Value | | | |
|---|---|---|---|---|---|---|---|---|---|---|
| | | | SIM-$Net_{CA}$ | | Δ | | SIM-$Net_{CA}$ | | Δ | |
| | Acc | F1 | Acc | F1 | Acc | F1 | Acc | F1 | Acc | F1 |
| *Thorns* | 95.20 | 93.47 | 94.00 | 94.05 | -1.20 | +0.58 | 95.00 | 94.99 | -0.20 | +1.52 |
| *Fruits* | 64.66 | 67.22 | 58.79 | 66.82 | -5.87 | -0.40 | 58.79 | 66.59 | -5.87 | -0.63 |
| *Leaves with acuminate tips* | 77.00 | 73.92 | 75.40 | 75.15 | -1.60 | +1.23 | 76.00 | 76.74 | -1.00 | +2.82 |
| *Infructescence* | 69.37 | 69.10 | 64.58 | 66.86 | -4.79 | -2.24 | 64.17 | 67.64 | -5.20 | -1.46 |
| *Leaves with an acute base* | 71.40 | 72.66 | 69.00 | 71.40 | -2.40 | -1.26 | 69.60 | 69.97 | -1.80 | -2.69 |

Table 9: Comparison of SIM-Net with simple concatenation (SIM-Net) versus SIM-Net with bidirectional cross-attention (SIM-$Net_{BCA}$), evaluated on two datasets: CIPD and SIPD.

| Trait | SIM-Net | | SIPD | | | | CIPD | | | |
|---|---|---|---|---|---|---|---|---|---|---|
| | | | SIM-$Net_{BCA}$ | | Δ | | SIM-$Net_{BCA}$ | | Δ | |
| | Acc | F1 | Acc | F1 | Acc | F1 | Acc | F1 | Acc | F1 |
| *Thorns* | 92.61 | 92.55 | 92.35 | 92.18 | -0.26 | -0.37 | 95.20 | 95.24 | +2.59 | +2.69 |
| *Fruits* | 64.88 | 67.86 | 60.42 | 67.03 | -4.46 | -0.83 | 62.24 | 66.75 | -2.64 | -1.11 |
| *Leaves with acuminate tips* | 81.82 | 81.40 | 72.06 | 73.42 | -9.76 | -7.98 | 76.40 | 76.89 | -5.42 | -4.51 |
| *Infructescence* | 66.35 | 69.51 | 62.83 | 67.47 | -3.52 | -2.04 | 65.83 | 69.41 | -0.52 | -0.10 |
| *Leaves with an acute base* | 73.87 | 74.92 | 69.53 | 73.68 | -4.34 | -1.24 | 70.00 | 69.49 | -3.87 | -5.43 |

## B. Ablation Studies

### B.1. Remove points from the point clouds

To investigate the impact of point density on the model performance, we conducted an ablation study where we randomly removed 60% and 90% of the points from the point clouds associated with the images. The results showed that the accuracy performance decreased as the number of points decreased (*cf.* Table 10), highlighting the importance of a sufficient number of points for better classification.

Table 10: Accuracy evaluation with ablation on the point clouds of the SIM-Net model on the selected unsegmented image point cloud dataset (SSIPD)

| Trait | 1024 points | 410 points (40%) | 102 points (10%) |
|---|---|---|---|
| Fruits | 66.56 | 62.01 | 61.64 |
| Leaves with acuminate tips | 79.84 | 76.47 | 71.47 |
| Infructescence | 70.19 | 66.08 | 64.01 |
| Leaves with an acute base | 74.67 | 70.97 | 69.18 |

### B.2. Point clouds without z coordinates

To assess the impact of integrating the RGB color information into the point cloud data on classification performance, we conducted an ablation study where the z-coordinate was set to zero in the four herbarium datasets (SPID, CIPD, SSIPD and CSIPD), effectively removing the depth information. We then compared the results to those obtained using point clouds where the z-coordinate was calculated as the average of the color intensities. The results, as illustrated in Table 11, and Table 12, demonstrates a significant decrease in accuracy and F1-score performance when the z-coordinate was excluded, with differences of up to 5.29% in accuracy on the SIPD dataset, 3.60% on the CIPD dataset, 3.98% on the SSIPD dataset, and 7.7% on the CSIPD dataset. This highlights the importance of depth information in capturing geometric relationships and improving robustness to variations in specimen color and preservation.

*Table 11: Evaluation of SIM-Net model performance on point clouds with and without z coordinate using unsegmented images.*

| Trait | Selected Image Point cloud (SIPD) | | | | | | Complete Image Point cloud (CIPD) | | | | | |
|---|---|---|---|---|---|---|---|---|---|---|---|---|
| | with z | | without z | | Δ | | with z | | without z | | Δ | |
| | Acc | F1 | Acc | F1 | Acc | F1 | Acc | F1 | Acc | F1 | Acc | F1 |
| **Thorns** | 95.75 | 95.74 | 94.20 | 94.05 | +1.55 | +1.69 | 95.00 | 95.01 | 91.60 | 91.49 | +3.40 | +3.52 |
| **Fruits** | 66.56 | 69.65 | 64.12 | 67.88 | +2.44 | +1.77 | 64.34 | 69.52 | 62.24 | 68.26 | +2.10 | +1.26 |
| **Leaves with acuminate tips** | 79.84 | 79.19 | 77.94 | 77.63 | +1.90 | +1.56 | 81.40 | 81.44 | 77.80 | 77.35 | +3.60 | +4.09 |
| **Infructescence** | 70.19 | 71.35 | 64.90 | 67.90 | +5.29 | +3.45 | 73.05 | 74.24 | 71.67 | 72.69 | +1.38 | +1.55 |
| **Leaves with an acute base** | 74.67 | 75.84 | 69.89 | 72.39 | +4.78 | +3.45 | 71.80 | 73.46 | 69.60 | 71.50 | +2.20 | +1.96 |

*Table 12: Evaluation of SIM-Net model performance on point clouds with and without z coordinate using segmented images.*

| Trait | Selected Segmented Image Point cloud (SSIPD) | | | | | | Complete Segmented Image Point cloud (CSIPD) | | | | | |
|---|---|---|---|---|---|---|---|---|---|---|---|---|
| | with z | | without z | | Δ | | with z | | without z | | Δ | |
| | Acc | F1 | Acc | F1 | Acc | F1 | Acc | F1 | Acc | F1 | Acc | F1 |
| **Thorns** | 92.61 | 92.55 | 90.24 | 90.49 | +2.37 | +2.06 | 95.90 | 95.79 | 88.20 | 88.80 | +7.7 | +6.99 |
| **Fruits** | 64.88 | 67.86 | 62.27 | 65.52 | +2.61 | +2.34 | 64.84 | 67.03 | 59.48 | 66.44 | +5.36 | +0.59 |
| **Leaves with acuminate tips** | 81.82 | 81.40 | 79.29 | 79.27 | +2.53 | +2.13 | 82.27 | 82.33 | 80.80 | 80.42 | +1.47 | +1.91 |
| **Infructescence** | 66.35 | 69.51 | 64.01 | 67.75 | +2.34 | +1.76 | 70.20 | 72.32 | 69.79 | 71.63 | +0.41 | +0.69 |
| **Leaves with an acute base** | 73.87 | 74.92 | 69.89 | 73.75 | +3.98 | +1.17 | 71.84 | 73.56 | 71.16 | 69.97 | +0.68 | +3.59 |

### C. Color encoding of pixels coordinates

To demonstrate the importance of explicitly incorporating x, y, z coordinates in the form of point clouds and processing them with PointNet-like architectures, we encoded the same spatial information into images referred to as Color-Coded Coordinates Masks (*cf.* Figure 5). We used then these masks in a dual ResNet101 encoder architecture, that we call, a Color-coded Mask and Image Fusion Network (CMI-FNet).

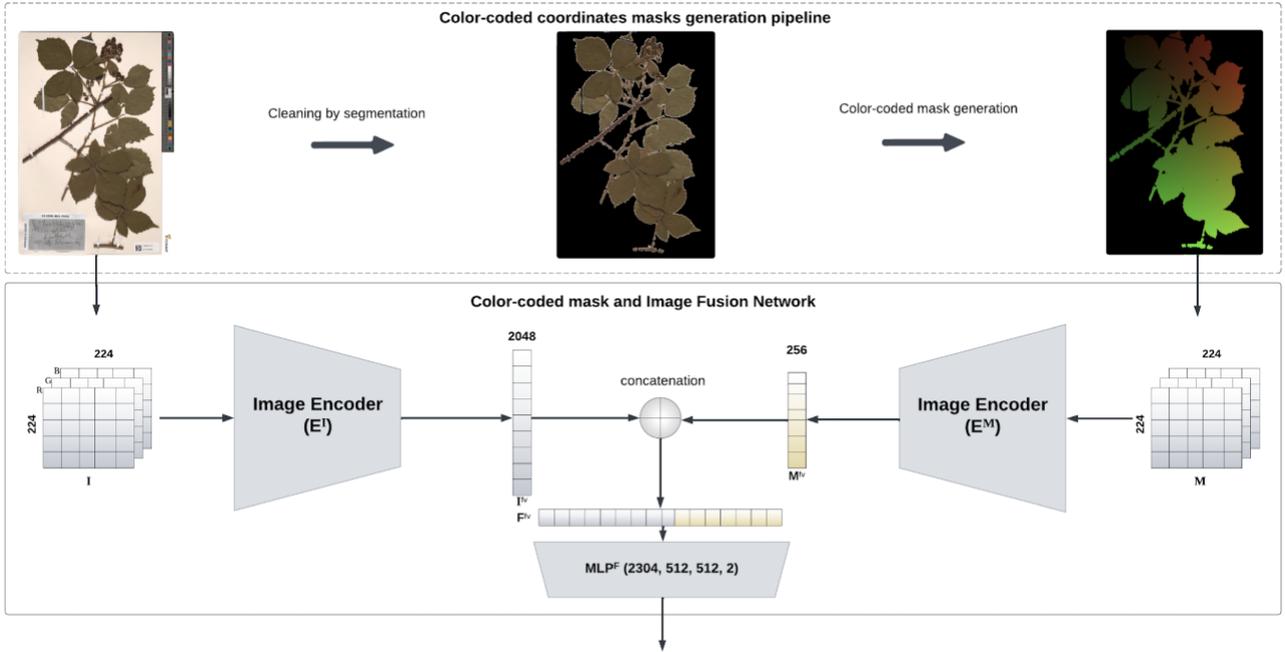

*Figure 5: Color-coded coordinates masks generation pipeline and the Color-coded Mask and Image Fusion Network (CMI-FNet).*

## C.1.  Color-coded coordinates masks generation pipeline

The transformation pipeline of herbarium sheet images into color-coded coordinates masks, as illustrated in Figure 5, begins with a segmentation phase that removes the background elements. The obtained image $I \in R^{H \times W \times C}$ is then transformed into another image $I'$, termed "Color-coded coordinate masks". This transformation is achieved by converting each pixel $I(x,y)$ with $(R_{xy}, G_{xy}, B_{xy}) \neq (0,0,0)$ to a new pixel $I'(x,y)$ with $(R'_{xy} = x, G'_{xy} = y, B'_{xy} = \frac{R_{xy}+G_{xy}+B_{xy}}{3})$. The values $(R'_{xy}, G'_{xy}, B'_{xy})$ are normalized to the interval [0, 256], assuming the pixel colors values $R'G'B'$ are scaled appropriately within this range.

## C.2.  Fusing RGB features with Color-coded masks features

The fusion function involves concatenating the RGB image feature vector, of size 2048, with the color-coded coordinates mask feature vector, of size 256 (*cf.* Figure 5). In this network, we use ResNet101 as the image encoder. To train the model, we used the same parameters as in SIM-Net training, except for the batch size, which we set to 64. The Figure 6 shows examples of herbarium images, their masks, color-coded masks, and the associated point clouds, respectively.

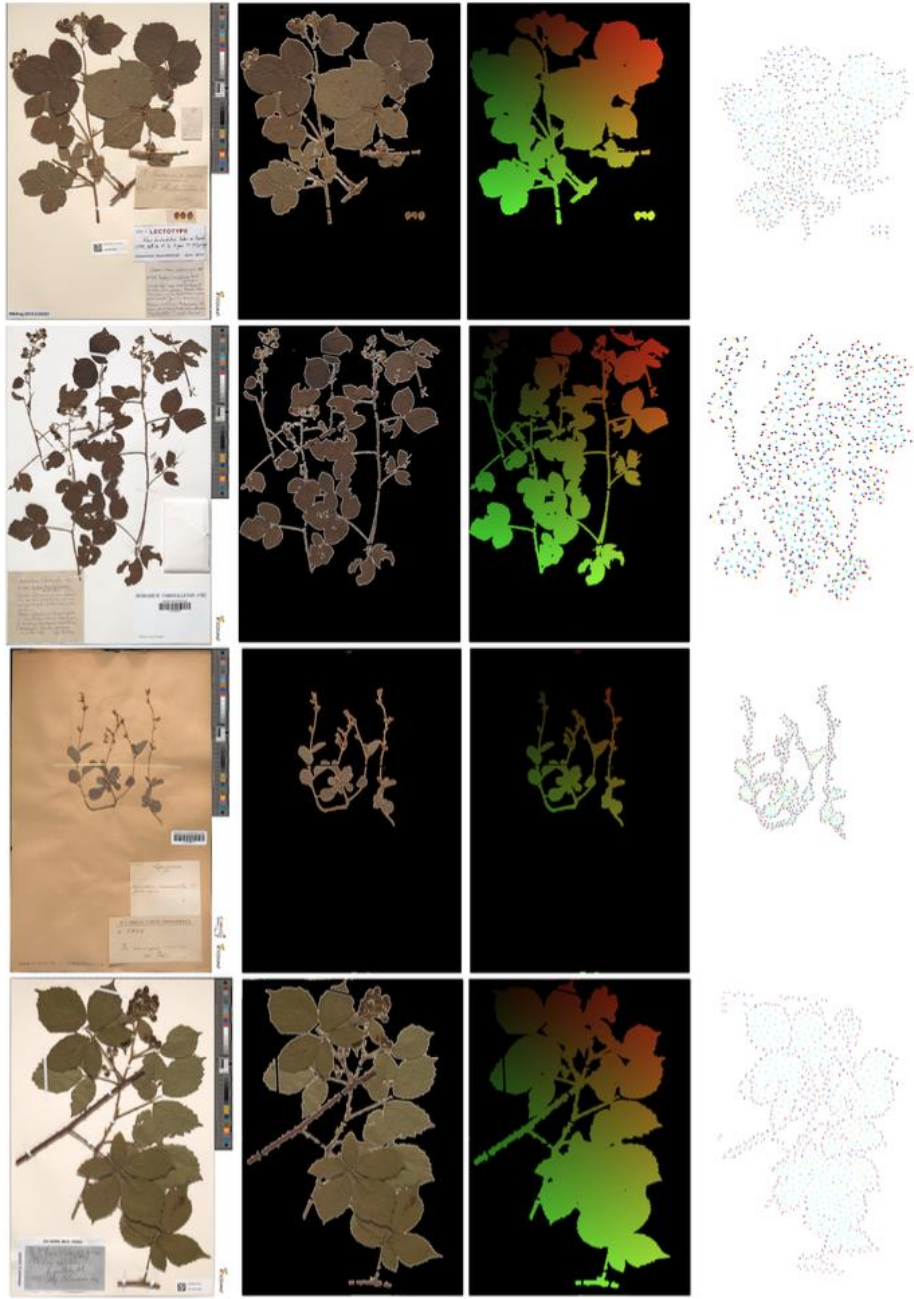

*Figure 6: Examples of herbarium scans with corresponding masks, color-coded masks, and point clouds.*

*Table 13: Performance evaluation of the SIM-Net model compared to a Color-coded Mask and Image Fusion Network (CMI-FNet) using pairs of (unsegmented images, color-coded masks) applied on the -- Selected Image Point Cloud Dataset (SIPD).*

| Trait | SIM-Net | | CCM-IFNet | | Δ | |
|---|---|---|---|---|---|---|
| | Acc | F1 | Acc | F1 | Acc | F1 |
| **Thorns** | 95.75 | 95.74 | 93.67 | 93.51 | +2.08 | +2.23 |
| **Fruits** | 66.56 | 69.65 | 62.27 | 66.78 | +4.29 | +2.87 |
| **Leaves with acuminate tips** | 79.84 | 79.19 | 74.71 | 75.60 | +5.13 | +3.59 |
| **Infructescence** | 70.19 | 71.35 | 67.55 | 70.44 | +2.64 | +0.91 |
| **Leaves with an acute base** | 74.67 | 75.84 | 70.25 | 74.12 | +4.42 | +1.72 |

We compared the results to our original architecture SIM-Net and we observed that the performance of the SIM-Net model is better than those obtained with the Color-coded coordinates mask and image fusion network, with improvements of up to more than 5% for accuracy and more than 3.5% for the F1-score (*cf.* Table 13).

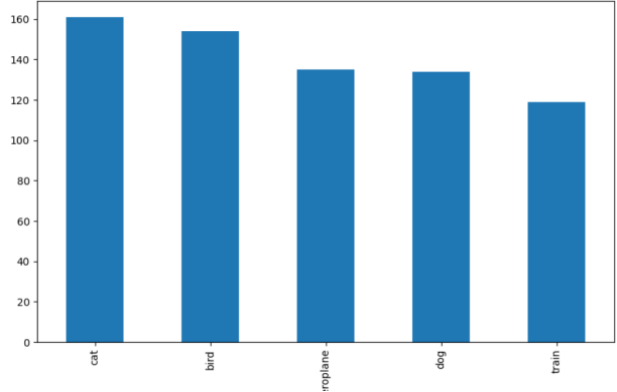

*Figure 7: Distribution of the 5 selected classes in the prepared VOC dataset.*

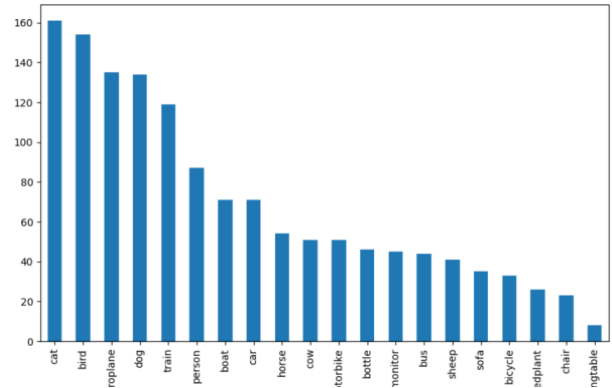

*Figure 8: Distribution of the 20 classes in the prepared VOC dataset.*

### D. Generalizability

To study the performance of our model on more general datasets with varied labels, we used the public dataset -- The PASCAL Visual Object Classes (PASCAL VOC)[3]. The dataset contains 2,913 images with their associated masks. However, some images contain multiple segmented objects. Consequently, using this dataset as it is, does not suit our study case in this paper since, as mentioned, we are interested in images targeting a single object. Therefore, we selected a subset of images containing a single object of interest and constructed a dataset of 1,389 images containing a single segmented object. The dataset is labeled with 20 different classes. We created two datasets. The first dataset, with 5 classes ('train', 'dog', 'cat', 'bird', 'airplane'), contains 700 images with a balanced distribution across the classes (Figure 7). The second dataset, with 20 classes, includes the 1,389 selected images but has a significant imbalance between the classes (Figure 8). For each image, we used its mask to generate the equivalent point clouds. Figure 9 shows examples of images and their masks from VOC dataset, with their corresponding generated point clouds.

*Table 14: Accuracy evaluation on VOC-Selected Dataset.*

|  | N° Classes | ResNet101 | SIM-Net |
|---|---|---|---|
| **ResNet101 from scratch** | 5 | 35.51 | 47.83 |
|  | 20 | 17.52 | 22.63 |
| **ResNet101 pretrained** | 5 | 90.38 | 70.62 |
|  | 20 | 39.78 | 24.45 |

Table 14 demonstrates that our method is more effective when we utilize ResNet101 built from scratch in both architectures. However, when ResNet101 is pre-trained on ImageNet 1K, our approach significantly underperforms compared to using ResNet101 alone, with discrepancies reaching up to 20%. This performance gap is likely influenced by the fact that our method focuses on classifying objects of a similar nature, such as herbarium images, whereas the VOC dataset comprises objects of a vastly different nature. Furthermore, in the VOC dataset, target objects are often partially occluded or incomplete, with segmentation results producing object shapes that do not accurately reflect their real-world geometry—unlike herbarium specimens, which are typically fully visible and free from such occlusions. Additionally, the close similarity between the VOC and ImageNet datasets may contribute to significant variances when generalizing results.

---

[3] http://host.robots.ox.ac.uk/pascal/VOC/.

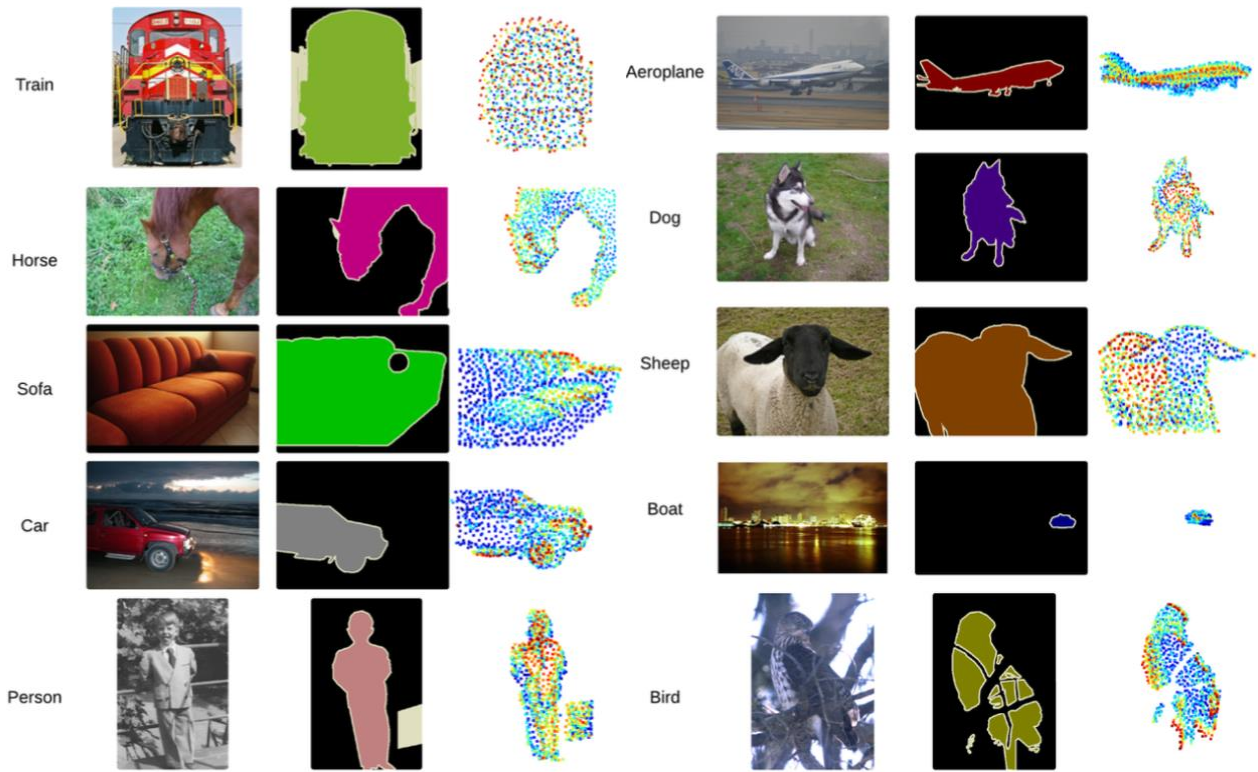

*Figure 9: Examples of images and their masks from VOC dataset, with their corresponding generated point clouds.*